\documentclass[11pt]{article}

\usepackage{amssymb}
\usepackage{amsmath,amsfonts,amsthm}
\usepackage{latexsym}
\usepackage{mathrsfs}
\usepackage{color}
\usepackage{dsfont}
\usepackage[overload]{empheq}
\usepackage{authblk}
\usepackage{hyperref}       
\hypersetup{
	colorlinks=true,
	linkcolor=blue,
	filecolor=magenta,
	urlcolor=cyan,
}
\usepackage{url}            
\usepackage{booktabs}       

\usepackage[top=1in, bottom=1in, left=1in, right=1in]{geometry}
\usepackage{bm}
\usepackage{graphicx}
\usepackage[numbers]{natbib}
\usepackage{makecell}
\usepackage{bigints}
\usepackage{tabularx}
\usepackage{pifont}
\usepackage{enumitem}
\usepackage{graphics}
\usepackage{algorithm,algorithmic}
\usepackage[toc, page]{appendix}
\usepackage{multirow}
\DeclareMathOperator*{\argmin}{arg\,min}
\DeclareMathOperator*{\argmax}{arg\,max}
\newtheorem{claim}{Claim}
\newcommand{\btheta}{\boldsymbol{\theta}}
\newcommand{\bs}{\boldsymbol}

\begin{document}
\title{Federated Learning with Uncertainty and Personalization via Efficient Second-order Optimization}

\author[1]{Shivam Pal}
\author[1]{Aishwarya Gupta}
\author[1]{Saqib Sarwar}
\author[1]{Piyush Rai}

\affil[1]{Department of Computer Science and Engineering, IIT Kanpur, India}
\affil[1 ]{\texttt {\{pshivam, aishwaryag, saqib, piyush\}@cse.iitk.ac.in}}
\date{}
\maketitle

\begin{abstract}
  Federated Learning (FL) has emerged as a promising method to collaboratively learn from decentralized and heterogeneous data available at different clients without the requirement of data ever leaving the clients. Recent works on FL have advocated taking a Bayesian approach to FL as it offers a principled way to account for the model and predictive uncertainty by learning a posterior distribution for the client and/or server models. Moreover, Bayesian FL also naturally enables personalization in FL to handle data heterogeneity across the different clients by having each client learn its own distinct personalized model. In particular, the hierarchical Bayesian approach enables all the clients to learn their personalized models while also taking into account the commonalities via a prior distribution provided by the server. However, despite their promise, Bayesian approaches for FL can be computationally expensive and can have high communication costs as well because of the requirement of computing and sending the posterior distributions. We present a novel Bayesian FL method using an efficient second-order optimization approach, with a computational cost that is similar to first-order optimization methods like Adam, but also provides the various benefits of the Bayesian approach for FL (e.g., uncertainty, personalization), while also being significantly more efficient and accurate than SOTA Bayesian FL methods (both for standard as well as personalized FL settings). Our method achieves improved predictive accuracies as well as better uncertainty estimates as compared to the baselines which include both optimization based as well as Bayesian FL methods.
\end{abstract}


\newpage


\section{Introduction}

Federated Learning (FL)~\citep{mcmahan2017communication} aims at learning a global model collaboratively across clients without compromising their privacy. It involves multiple client-server communication rounds, where in each round the selected clients send their local models (trained on their private dataset) to the server and the server aggregates the received models followed by its broadcasting to all clients. Thus, the global model, an approximation to the model obtained if all the data was accessible, depends significantly both on the quality of the received clients' models and the chosen aggregation strategy at the server. As a result, a straightforward approach like FedAvg\citep{mcmahan2017communication} can yield a high-performing global model if the data is i.i.d. distributed among clients; however performs suboptimally in case of non i.i.d. data distribution. Moreover, the challenges are compounded if each client has a limited private dataset.

The limitations of standard FL become even more apparent with data heterogeneity, where clients have distinct data distributions. A single global model might fail to represent all clients well, leading to poor performance. This motivates personalized FL (pFL)~\cite{tan2022towards}, which aims to adopt models to individual clients while leveraging shared global knowledge. 


In such settings, learning the posterior \emph{distribution} instead of a point estimate at each client results in enhanced performance and uncertainty measures, as demonstrated in several recent works, such as~\citep{al2020federated,Liu_2024,bhatt2023federatedlearninguncertaintydistilled,guo2023federated} which have advocated taking a Bayesian approach to FL. Moreover, Bayesian FL is also natural for personalization because the server model can serve as a prior distribution in a hierarchical Bayesian framework, enabling easy personalization of client models using their respective client-specific likelihoods.
  However, existing Bayesian FL and pFL methods usually rely on running computationally expensive routines on the clients (e.g., requiring expensive MCMC sampling~\citep{al2020federated}, expensive Laplace's approximation which requires Hessian computations~\citep{Liu_2024} on the clients, or methods based on learning deep ensembles~\citep{linsner2021approaches}), as well as expensive client-server communication~\citep{kassab2022federated} and aggregation at the server (note that, unlike standard FL, Bayesian FL would require sending the whole client posterior to the server). Due to such computational bottlenecks and communication overhead, Bayesian approaches lack scalability, especially for clients with limited resources and bandwidth.


\begin{figure*}[!htbp]
    \centering
    \includegraphics[width=1\textwidth]{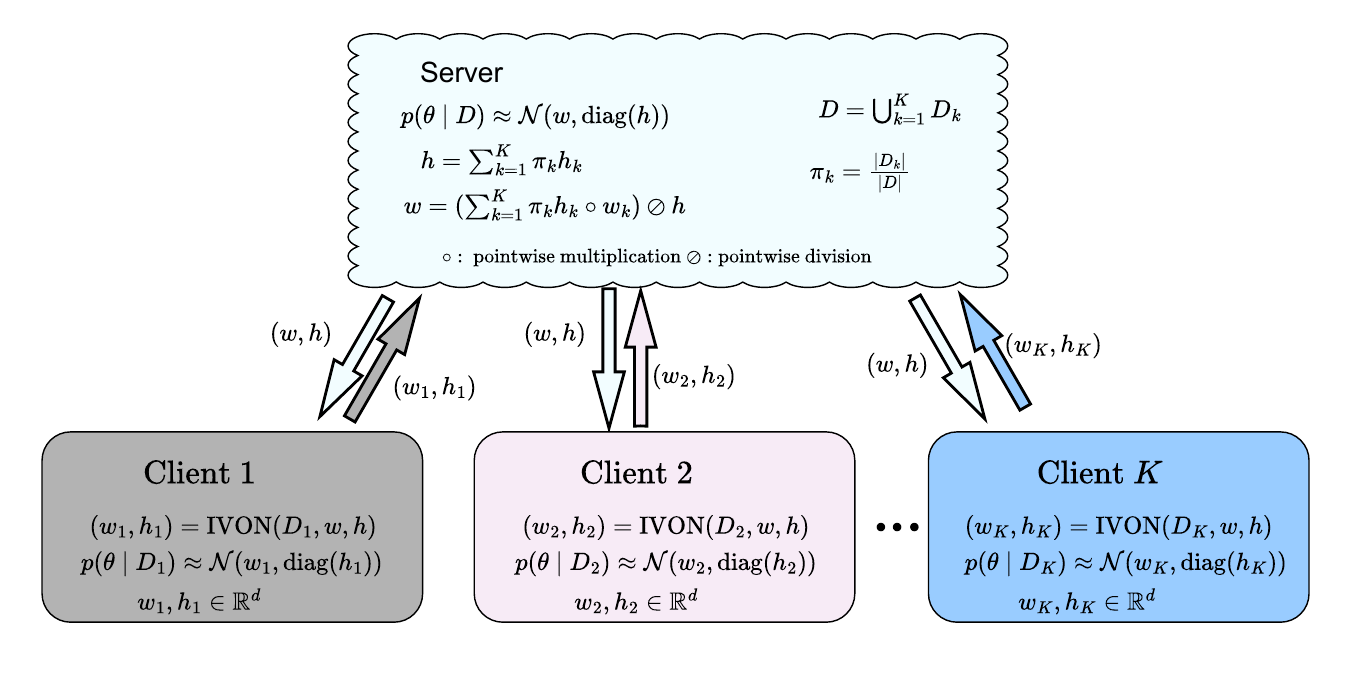}
    \caption{Illustration of FedIVON. }
    \label{fig:main_diagram}
\end{figure*}

Thus, 
to bridge this gap, we propose a novel Bayesian FL algorithm FedIvon (with its high-level idea illustrated in Fig.~\ref{fig:main_diagram}), that balances the benefits of Bayesian inference - enhanced performance, and quantification of predictive uncertainty - with minimal increase in computational and communication overhead. 
In particular, we leverage the IVON (Improved Variational Online Newton) algorithm \citep{shen2024variationallearningeffectivelarge} to perform highly efficient variational inference (VI) on each client by approximating its local posterior using a Gaussian with diagonal covariance. It uses the natural gradient to capture the geometry of the loss function for faster convergence. Moreover, it computes the Hessian implicitly, making our method computationally cheaper than other existing Bayesian FL and pFL methods that use explicit Hessian computation, e.g., Laplace's approximation~\citep{Liu_2024}, expensive MCMC sampling~\citep{al2020federated,bhatt2023federatedlearninguncertaintydistilled}, or even VI~\citep{kassab2022federated} at the clients. These local posteriors can be efficiently sent to the server and the global posterior can be computed for which we also present local posterior aggregation strategies. 
Our main contributions are:
\begin{itemize}
    \item We introduce a Bayesian FL method FedIvon that uses an efficient second-order optimization approach for variational inference, maintaining computational costs similar to first-order methods like Adam.
    \item Our method demonstrates improvements in predictive accuracy and uncertainty estimation compared to state-of-the-art (SOTA) Bayesian and non-Bayesian FL methods.
    \item Our method also supports client-level model personalization naturally by leveraging a hierarchical Bayesian framework. Clients can use the server's posterior as priors to learn their private models, effectively balancing local adaptation with global knowledge sharing.
\end{itemize}

\section{Related Work}
FedAvg~\cite{mcmahan2017communication}, the foundational federated learning algorithm, approximates the global model as the weighted aggregation of locally trained client models, performing effectively with i.i.d. data distributions. Since then, numerous sophisticated and efficient algorithms have been proposed to handle more realistic challenges such as non-i.i.d. data distribution, heterogeneous and resource-constrained clients, and multi-modal data as explored in recent survey works~\cite{lin2020ensemble, Pfeiffer2023FederatedLF, Zhang2023ASO, Che2023MultimodalFL, Liu2023RecentAO}. However, here, we will restrict our discussion to Bayesian FL and personalized FL algorithms as they are most relevant to our work. 

\textbf{Bayesian Federated Learning} A key limitation of point-estimate-based approaches is their susceptibility to overfitting in limited data settings and lack of predictive uncertainty estimates.  To address this, Bayesian approaches have been advocated for federated learning, which involves the computation of clients' local posterior distribution followed by their aggregation at the server to compute the global posterior distribution, offering enhanced performance and quantification of predictive uncertainty. Unfortunately computing full posterior distribution is intractable and poses communication overhead. FedBE~\cite{DBLP:journals/corr/abs-2009-01974} mitigates the communication overhead by leveraging SWAG~\cite{maddox2019simple} to learn each client's posterior but communicating only its mean. The server then fits a Gaussian/Dirichlet distribution to the clients' posterior mean and distills it into a single model to be communicated in the next round. However, FedBE does not incorporate clients' covariances, omitting the underlying uncertainty in their models during aggregation. FedPA~\cite{al2020federated} addresses this by learning a Gaussian distribution for each client and computes the mean of the global posterior at the server. However, it eventually discards the covariance of the global posterior and computes a point estimate to limit the communication costs. Similarly, FedLaplace~\cite{Liu_2024} approximates each client's posterior as a Gaussian distribution, modeling the global posterior as a mixture of Gaussian, though eventually it too simplifies it to a single Gaussian by minimizing KL divergence.

\textbf{Second-order Optimization for Federated Learning} shows promise for improving convergence but is often limited by efficiency and communication overhead. Methods such as FedNL~\cite{Safaryan2021FedNLMN}, which use privacy-preserving Hessian learning and compression, and second-order approaches incorporating global line search~\cite{bischoff2021secondorderoptimizationmethodsfederated}, offer potential solutions to these challenges.

\textbf{Personalized Federated Learning} In the case of non-iid data distribution among clients, a single global model represents the average data distribution and diverges substantially from each client's local distribution. Consequently, the global model, though benefitted from collaborative learning, performs suboptimally for individual clients. Personalized federated learning addresses this challenge by adapting a part or the whole model to the local data distribution explicitly. A typical approach is to split the model into two parts - a base model for global representation learning and a head model for personalized learning. FedPer~\cite{achituve2021personalized} and FedRep~\cite{collins2021exploiting} use this strategy, applying FedAvg for collaborative learning of the base model leveraged by the head for local data adaptation. Similarly, FedLG~\cite{liang2020think} splits the model into local and global components to learn local and shared representations respectively. It shares the global parameters with the server while enhancing local parameters further using the unsupervised or self-supervised approach. PerFedAvg~\cite{fallah2020personalized} applies a Model-Agnostic Meta-Learning (MAML)~\cite{finn2017model} inspired framework to learn a shared model for faster adaptation to the client's data. pFedME~\cite{t2020personalized} decouples personalized adaptation from shared learning by regularizing each client's loss function using Moreau envelopes. pFedBayes~\cite{zhang2022personalized} is a Bayesian approach that aims at learning the personalized posterior distribution of each client. In each round, pFedBayes computes clients' posterior using the global model as the prior and sends it to the server for updating the global model. pFedVEM~\cite{zhu2023confidence} also computes the client's posterior by restricting it to the Gaussian family. However, it leverages the collaborative knowledge of other clients by assuming conditional independence among clients' models given the global model.






\section{Bayesian FL via Improved Variational Online Newton}

The standard formulation of FL is similar to distributed optimization except some additional constraints, such as no data sharing among clients and server and a limited communication budget. Assuming $K$ clients, let $\mathcal{D} = \bigcup_{k \in [K]} \mathcal{D}_k$ be the total available data where $\mathcal{D}_k$ denotes the private data of client $k$. The objective of standard FL is to solve $\boldsymbol{\theta}^{*} = \argmin_{ \boldsymbol{\theta}} \sum_{k \in [K]} - \log p(\mathcal{D}_k \mid \boldsymbol{\theta} )$. However, this optimization problem is not trivial as it requires access to each client's data which is not permitted in the federated setting. Thus, a multi-round approach is usually taken where clients learn their local models,  send these local models to a central server which aggregates them into a global model, and send the global model to the clients to continue the next round of learning.

Unlike standard FL which only learns a point estimate of $\boldsymbol{\theta}$, an alternative is to learn a \emph{distribution} of $\boldsymbol{\theta}$. The posterior distribution of $\boldsymbol{\theta}$ can be written as 
\begin{equation}
p(\boldsymbol{\theta} \mid \mathcal{D}) \propto p(\boldsymbol{\theta}) \prod_{k \in [K]} p(\mathcal{D}_k \mid \boldsymbol{\theta})
\end{equation}
where $p(\boldsymbol{\theta})$ is prior distribution on $\boldsymbol{\theta}$ and $p(\mathcal{D}_k \mid \boldsymbol{\theta})$ is data likelihood of client $k$. Assuming uniform prior $p(\boldsymbol{\theta})$, it can be trivially shown that optimizing the standard FL objective function is equivalent to finding the mode of the posterior $p(\boldsymbol{\theta} \mid \mathcal{D})$, i.e., $\boldsymbol{\theta}^* = \argmax_{ \boldsymbol{\theta}} \ \log p(\boldsymbol{\theta} \mid \mathcal{D})$.

Computing the full posterior  $p(\boldsymbol{\theta} \mid \mathcal{D})$ is more useful than computing just the point estimate $\boldsymbol{\theta}^*$ because the posterior helps take into account model uncertainty. However, it is computationally intractable to compute the posterior exactly. Directly approximating $p(\boldsymbol{\theta} \mid \mathcal{D})$ using approximate inference methods such as MCMC or variational inference~\citep{angelino2016patterns} is also non-trivial, as it requires computing each client's likelihood which in turn requires global access to all the client's data. 
\begin{claim}
    The global posterior $p(\boldsymbol{\theta} \mid \mathcal{D})$ can be approximated at the server by the product of local client posteriors without requiring access to any client's local data.
\end{claim}
If local posteriors $p(\theta \mid \mathcal{D}_k)$ are also being approximated, multiple rounds of optimization are needed to reduce the aggregation error in the global posterior~\citep{al2020federated}. In FL, another challenge is to make the computation of the local posteriors, their aggregation at the server, and the client-server communication, efficient, which in general can be difficult even for simple models~\citep{al2020federated}.

\subsection{Client's posterior approximation}
Assuming client $k$ has $N_k$ training examples, its local loss can be defined as $\bar{\ell}_k(\btheta) = \frac{1}{N_k} \sum_{i=1}^{N_k} \ell_i(\btheta)$, and we can compute the point estimate of the parameters as $\btheta^*_k = \argmin_{\btheta} \ \bar{\ell}_k(\btheta)$.
However, in our Bayesian FL setting, we will compute the (approximate) posterior distribution for each client using variational inference, which amounts to solving the following optimization problem
\begin{align}
    q^*_k(\btheta) &= \argmin_{q_k(\btheta)} \mathcal{L}_k(q)  \\   \mathcal{L}_k(q) &= \mathbb{E}_{q_k(\btheta)}[\bar{\ell}_k(\btheta)] + \mathbb{D}_{KL}(q_k(\btheta) \| p_k(\btheta))
    \label{eq:client-vi}
\end{align}
where $p_k(\btheta)$ is the prior and $\mathbb{D}_{KL}$ is the
Kullback-Leibler divergence. If we use the Gaussian variational family for $q_k(\btheta)$ with diagonal covariance then  $q_k(\btheta) =  
\mathcal{N}(\btheta | \boldsymbol{m}_k, \text{diag}(\boldsymbol{\sigma}_k^2))
 $, where $\boldsymbol{m}_k$ and $\boldsymbol{\sigma}_k^2$ denote the variational parameters that are to be optimized for. Optimizing the objective in Equation~\ref{eq:client-vi} w.r.t these variational parameters requires making the following updates
\begin{align}
\bs{m}^{t+1}_k &= \bs{m}^{t}_k - \alpha \Hat{\nabla}_{\bs{m}_k} \mathcal{L}_k(q) \\
\bs{\sigma}_k^{t+1} &= \bs{\sigma}_k^{t} - \alpha \Hat{\nabla}_{\bs{\sigma}_k} \mathcal{L}_k(q)
\label{eq:update_eq}
\end{align}
where $\alpha > 0$ is the learning rate.\\
Computing exact gradients in the above update equations is difficult due to the expectation term in $\mathcal{L}_k(q)$. A na\"ive way to optimize is to use stochastic gradient estimators. However, these approaches are not very scalable due to the high variance in the gradient estimates. 
\citep{shen2024variationallearningeffectivelarge} improved these update equations and provided much more efficient update equations similar to Adam optimizer, which is essentially the improved variational online Newton (IVON) algorithm~\citep{shen2024variationallearningeffectivelarge}, with almost exact computational cost as Adam, and their key differences are summarized below
\begin{itemize}
    \item Unlike Adam which solves for $\bs \theta$, IVON solves for both the mean vector $\bs{m}$ and the variances $\bs{\sigma}^2$ which provides us an estimate of the Gaussian variational approximation at each client. Note that the mean $\bs{m}$ plays the role of $\bs \theta$ in Adam. In addition, the variances naturally provide the uncertainty estimates for $\bs \theta$, essential for Bayesian FL (both in estimating the client models' uncertainties as well as during the aggregation of client models at the server).
    \item Unlike Adam which uses squared minibatch gradients to adjust the learning rates in different dimensions, IVON uses a reparametrization defined as gradient element-wise multiplied by $(\bs \theta-\mathbf{m})/\bs \sigma^2$ to get an unbiased estimate of the (diagonal) Hessian. Using this, IVON is able to get a cheap estimate of the Hessian, which makes it a second-order method, unlike Adam.
    \item IVON offers the significant advantage of providing an estimate of second-order information \( h \) with minimal computational overhead. The Hessian \( h \) corresponds to the inverse of \(\sigma^2\), where \(\sigma^2 = \frac{1}{h + \delta}\). An estimate of \( h \) is accessible throughout the training process (see Algorithm \ref{algo:client}). Moreover, there is no explicit update question for \(h \). It is computed implicitly using gradient information. In comparison, standard optimization methods such as SGD, Adam, and SWAG require additional effort to estimate second-order information.
\end{itemize}

\begin{algorithm}[!htbp]
    \caption{FedIvon Algorithm}
    \begin{algorithmic}[1]
        \STATE \textbf{Input:} Total communication rounds $R$, total clients $K$, clients' private datasets $\{\mathcal{D}_k\}_{k=1}^{K}$, initial model weight $\boldsymbol{\tilde{m}_0}$, initial model Hessian $\boldsymbol{\tilde{h}_0}$

        \vspace{0.2cm}
        \FOR{$r = 1$ to $R$}
            \STATE Broadcast $\boldsymbol{\tilde{m}_r}, \boldsymbol{\tilde{h}_r}$ to all $K$ clients
            
            \vspace{0.2cm}
            \STATE Randomly sample $k$ clients
            \hfill\COMMENT{\textit{Update selected client models locally}}
            \FOR{$i =1$ to $k$}
                \STATE $\bs{m_i}, \bs{h_i} =$ Client\_Update($D_i, \boldsymbol{\tilde{m}_r}, \boldsymbol{\tilde{h}_r}$)
            \ENDFOR
            
            \vspace{0.2cm}
            \STATE Initialize $\boldsymbol{\tilde{m}_{r+1}} \leftarrow 0, \boldsymbol{\tilde{h}_{r+1}} \leftarrow 0$  \hfill\COMMENT{\textit{Aggregation of client models at server}}
            \FOR{$i = 1$ to $k$}
                \STATE $\boldsymbol{\tilde{h}_{r+1}} \leftarrow \boldsymbol{\tilde{h}_{r+1}} + \bs{h}_i * w[i]$ 
                \STATE $\boldsymbol{\tilde{m}_{r+1}} \leftarrow \boldsymbol{\tilde{m}_{r+1}} + \bs{m}_i \odot\bs{h}_i * w[i]$ 
            \ENDFOR

            \vspace{0.2cm}
            \STATE $\boldsymbol{\tilde{m}_{r+1}} \leftarrow \frac{\boldsymbol{\tilde{m}_{r+1}}}{\boldsymbol{\tilde{h}_{r+1}}}$
             (elementwise division)
            \hfill\COMMENT{\textit{Global weight and Hessian}}
        \ENDFOR

        \vspace{0.2cm}
        \STATE \textbf{Output:} Global model weights and Hessian $(\boldsymbol{\tilde{m}_{R}}, \boldsymbol{\tilde{h}_{R}})$
    \end{algorithmic}
\end{algorithm}

\begin{algorithm}[!htbp]
    \caption{Client\_Update}
    \label{algo:client}
    \begin{algorithmic}[1]
        \STATE \textbf{Input:} Local dataset $D$, model weights $\bs{m}$, Hessian($\bs{h}$), local\_epochs($E$), learning rates $\{\alpha_e\}$, weight decay $\delta$, hyperparameters $\beta_1, \beta_2$, batch-size $B$
        \STATE \textbf{Output:} Trained model weights $\bs{m}$, Hessian $\bs{\sigma}$

        \vspace{0.2cm}
            \STATE $\mathbf{g} \leftarrow 0, \quad \lambda \leftarrow |D|, n = E*|D|/B$.
            \STATE $\boldsymbol{\sigma} \leftarrow 1 / \sqrt{\lambda(\mathbf{h}+\delta)}$.
            \STATE $\alpha_e \leftarrow\left(h+\delta\right) \alpha_e$ for all $e \in \{1, 2, \dots, n\}$.
            \FOR{$e=1$ to $E$}
                \STATE Sample a batch of inputs of size $B$ from $D$.
                \STATE$\widehat{\mathrm{g}} \leftarrow \widehat{\nabla} \bar{\ell}(\boldsymbol{\theta})$, where $\boldsymbol{\theta} \sim q$
                \STATE$\widehat{\mathbf{h}} \leftarrow \widehat{\mathrm{g}} \cdot(\boldsymbol{\theta}-\mathbf{m}) / \bs{\sigma^2}$
                \STATE$\mathbf{g} \leftarrow \beta_1 \mathbf{g}+\left(1-\beta_1\right) \widehat{\mathbf{g}}$
                \STATE$\mathbf{h} \leftarrow \beta_2 \mathbf{h}+\left(1-\beta_2\right) \widehat{\mathbf{h}}+\frac{1}{2}\left(1-\beta_2\right)^2(\mathbf{h}-\widehat{\mathbf{h}})^2 /(\mathbf{h}+\delta)$
                \STATE$\overline{\mathbf{g}} \leftarrow \mathbf{g} /\left(1-\beta_1^e\right)$
                \STATE$\mathbf{m} \leftarrow \mathbf{m}-\alpha_e(\overline{\mathbf{g}}+\delta \mathbf{m}) /(\mathbf{h}+\delta)$
                \STATE$\boldsymbol{\sigma} \leftarrow 1 / \sqrt{\lambda(\mathbf{h}+\delta)}$
            \ENDFOR
        
    \end{algorithmic}
\end{algorithm}

\subsection{Posterior aggregation at server}
At the server, we can aggregate the client posteriors to compute the global posterior~\citep{fischer2024federated}. IVON approximates clients' posteriors as Gaussians and product of Gaussian distributions is still a Gaussian distribution up to a multiplicative constant. Thus we approximate the global distribution as a Gaussian whose optimal mean and covariance matrix expressions are given below. Moreover, since each client's variational approximation is a Gaussian with diagonal covariance matrix, it makes the aggregation operations efficient.
Let's assume  $q( \boldsymbol \theta \mid \mathcal{D}_k) = \mathcal{N}(\boldsymbol \theta \mid \boldsymbol \mu_k, \boldsymbol \Lambda_k^{-1})$ where $\boldsymbol \mu_k = \mathbf{m}_k$ and $\boldsymbol \Lambda_k = \text{diag}(\boldsymbol{\sigma}_k^2)$. Using results of the product of Gaussians based aggregation~\citep{Liu_2024,fischer2024federated}, we have
\begin{equation}
         \log q(\boldsymbol \theta \mid \mathcal{D}) \approx \sum_{k=1}^{K} w_k \log q(\boldsymbol \theta \mid \mathcal{D}_k)
\end{equation}         
where $w_k = \frac{N_k}{\sum_{k=1}^{K} N_k}$ and
\begin{equation}
    q(\boldsymbol \theta \mid \mathcal{D}) \approx \mathcal{N}(\boldsymbol \theta \mid \boldsymbol \mu, \boldsymbol \Lambda^{-1})
\end{equation}    
where $\boldsymbol \Lambda = \sum_{k=1}^{K} w_k \boldsymbol \Lambda_k$ and $\boldsymbol \mu = \boldsymbol \Lambda^{-1} \sum_{k=1}^{K} w_k \boldsymbol \Lambda_k \boldsymbol \mu_k$.

Other aggregation strategies are also possible~\citep{fischer2024federated} and we leave this for future work. Note that our aggregation strategy can also be seen as Fisher-weighted model merging~\citep{daheim2023model} where each client model is represented as the mean weights $\bs{m}_k$ and a Fisher matrix which depends on local posterior's variances $\bs{\sigma}_k^2$ (although model merging only computes the mean, not the covariance, and thus does not yield a global posterior distribution at the server).

The appendix provides further details of IVON and its integration in our Bayesian FL setup.

Notably, FedIvon is appealing from two perspectives: It can be viewed an an efficient Bayesian FL algorithm offering the various benefits of the Bayesian approach, as well as a federated learning algorithm that easily incorporates second-order information during the training of the client models, while not incurring the usual overheads of second-order methods used by some FL algorithms~\citep{bischoff2021second}.

\subsection{Personalized Federated Learning}
Personalized FL in FedIVON can be achieved straightforwardly. Similar to equation \ref{eq:client-vi},  the personalized loss function for each client $k$ is defined as,
\begin{equation}
    \mathcal{L}_k(q) =  \mathbb{ E}_{q_k(\btheta)}[\bar{\ell}_k(\btheta)] + \beta\; \mathbb{D}_{KL}(q_k(\btheta) \| p_k(\btheta)).
\end{equation}
Where $\beta \geq 0$ controls the level of personalization. The term $p_k(\btheta)$ represents the prior distribution for client $k$. During each communication round, the posterior distribution from the server can be used as the prior $p_k(\btheta)$ for the client. This setup enables clients to adapt the global model according to their local data characteristics while leveraging information from the global model.

When $\beta=0$, the model becomes fully personalized, relying solely on the client's data without influence from the prior (i.e., no information from the server). Conversely, a higher value of $\beta$  incorporates more knowledge from the global server model into the client's learning process, balancing between personalization and shared global information. This framework provides a flexible mechanism to adapt client models according to their individual data while still benefiting from collective learning through the shared server posterior. We fixed $\beta=1$ in all our pFL experiments.

\section{Experiments: Standard FL}

We experiment on three publicly available datasets: EMNIST \citep{Cohen2017EMNISTAE}, SVHN \citep{Netzer2011ReadingDI} and CIFAR-10 \citep{Krizhevsky2009LearningML}. EMNIST consists of 28x28 grayscale images of alphabets and digits (0-9) with a train and test split comprising $124800$ and $20800$ images respectively; however, in our experiments, we restrict to alphabets only.
SVHN consists of 32x32 RGB images of house number plates categorized into 10 distinct classes, each corresponding to one of the ten digits. It has a train and test split of size $73252$ and $26032$ respectively. 
CIFAR-10 comprises 32x32 RGB images of objects classified into 10 classes with $50000$ training images and $10000$ test images.

In our experiments, We use ADAM optimizer with \texttt{learning\_rate=1e-3, weight\_decay=2e-4} for FedAvg and FedLaplace method. IVON\cite{shen2024variationallearningeffectivelarge} optimizer is used for FedIvon with different hyperparameters given in Table \ref{tab:my_tab}. Linearly decaying learning rate is used in all the experiments.

\begin{table}[!htbp]
\scriptsize
    \centering
    \begin{tabular}{cccc}
        \toprule
        params & SVHN  & EMNIST & CIFAR-10 \\
        \midrule
         initial learning rate& 0.1 & 0.1 & 0.1\\
         final learning rate& 0.01 & 0.01 & 0.01 \\
         weight decay & 2e-4 & 2e-4 & 2e-4 \\
         batch size & 32 & 32 & 32\\
         ESS ($\lambda$)& 5000 & 5000 & 5000\\
         initial hessian ($h_0$)& 2.0 & 5.0 & 1.0\\
         MC sample while training& 1 &1  & 1\\
         MC samples while test & 500 & 500 & 500\\
         \bottomrule
    \end{tabular}
    \caption{Ivon Hyperparameters for FL experiments}
    \label{tab:my_tab}
\end{table}

We evaluate FedIvon in a  challenging and realistic scenario involving heterogeneous data distribution among a large number of clients with each client having very few training examples. 
For each experiment, we consider a total of $200$ clients with each client having a small private training set of less than $100$ examples. To simulate non-iid data distribution, we randomly sample inputs from the training split, partition the sampled inputs into shards, and distribute shards among clients to create class-imbalanced training data similar to \citep{DBLP:journals/corr/abs-2009-01974}. For a fair comparison, we use the same non-iid data split across clients for all the baseline methods and FedIvon.
We follow the experimental setup of~\citep{bhatt2023federatedlearninguncertaintydistilled} and train customized CNN models on EMNIST, SVHN, and CIFAR-10 datasets. We compare our proposed method FedIvon with FedAvg~\citep{mcmahan2017communication} (simple aggregation of client models at server) and FedLaplace~\citep{Liu_2024} (using the Laplace's approximation to fit a Gaussian distribution to each client's local model followed by aggregation at the server). FedAvg serves as a baseline to emphasize the importance of uncertainty quantification without compromising on the performance
while FedLaplace serves as a competitive baseline to evaluate FedIvon's predictive uncertainty measures. For all the baselines and FedIvon, we run the federated algorithm for $2000$ communication rounds, selecting a randomly sampled $5\%$ i.e., $10$ clients per round. We train each client's model locally for $2$ epochs using a batch size of $32$. We provide further details on hyperparameters, model architectures, and split in the appendix.

\subsection{Classification Task}

We train a classification model in FL setting using all the methods and report the results in Table~\ref{tab:result_table}. We evaluate all trained models' performance (accuracy and negative log-likelihood) on the test split and use metrics such as Expected Calibration Error (ECE) and Brier score to quantify predictive uncertainty. In our results, FedIvon@mean denotes point estimate based predictions evaluated at the mean of IVON posterior and FedIvon corresponds to Monte Carlo averaging with $500$ samples.

\begin{figure}[!htbp]
        \includegraphics[scale=0.4]{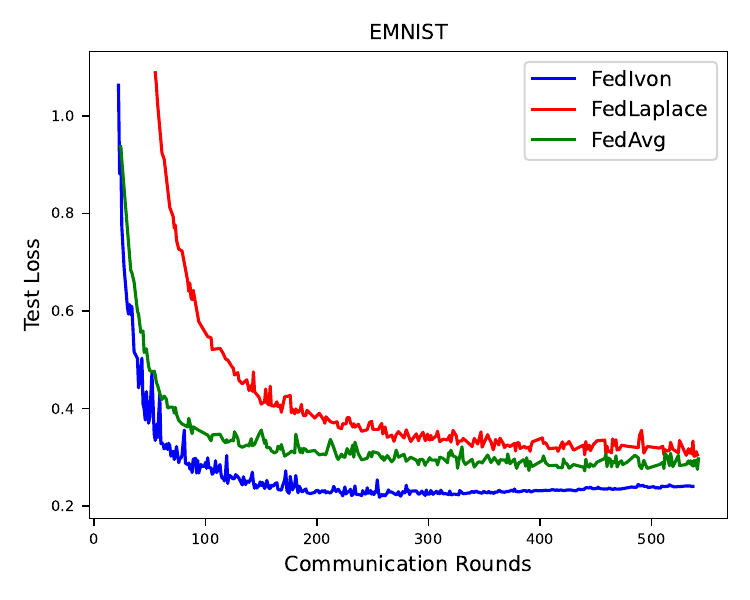}
        \includegraphics[scale=0.4]{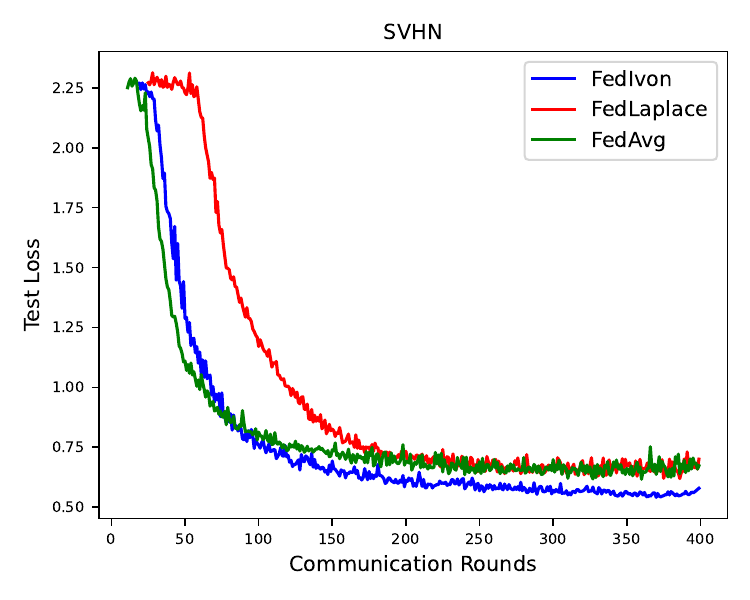}
        \includegraphics[scale=0.4]{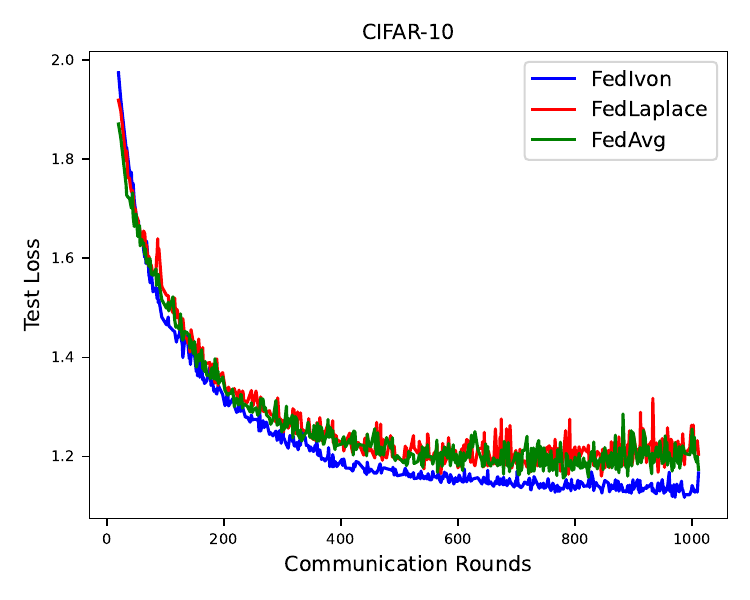}
    \caption{Loss of various methods vs rounds (left: EMNIST, center: SVHN, right: CIFAR-10).}
    \label{fig:loss_plot}
\end{figure}

\begin{figure}[!htbp]
        \includegraphics[scale=0.4]{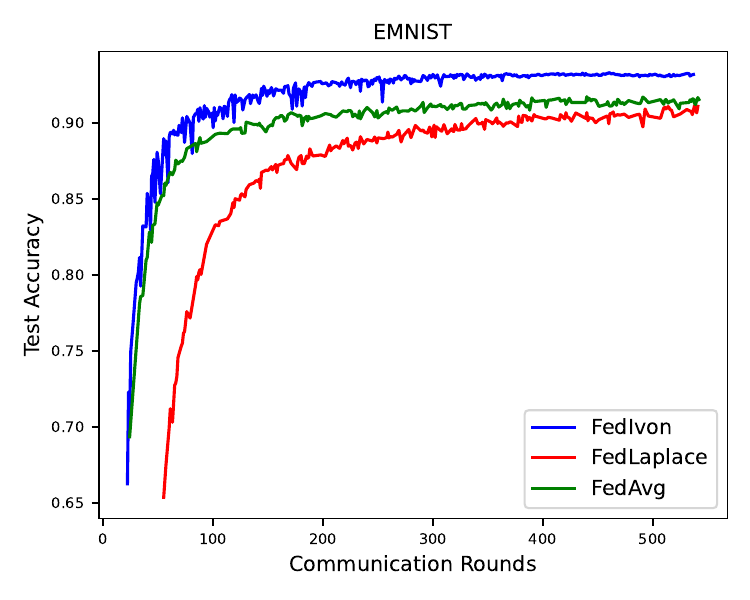}
        \includegraphics[scale=0.4]{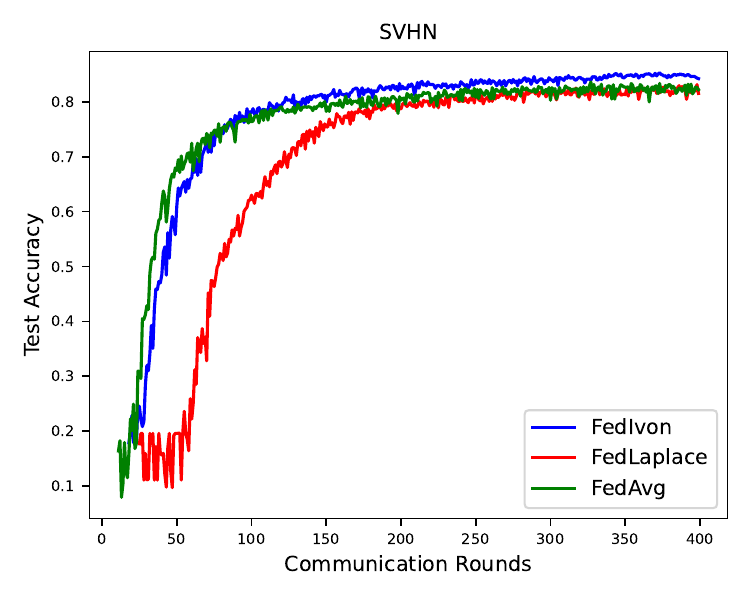}
        \includegraphics[scale=0.4]{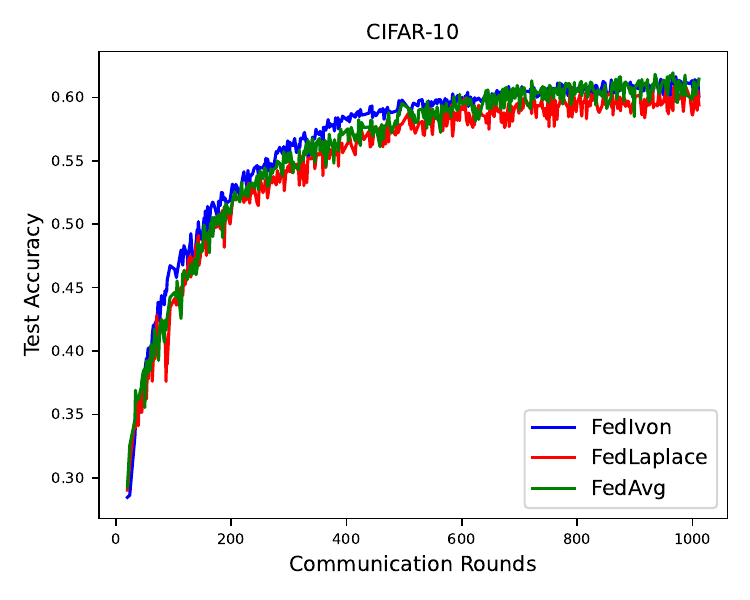}
    \caption{Test accuracy vs rounds (left: EMNIST, center: SVHN, right: CIFAR-10).}
    \label{fig:acc_plot}
\end{figure}

As shown in Table~\ref{tab:result_table}, FedIvon outperforms all the baselines and yields the best test performance and calibration scores. FedIvon leverages the improved variational online Newton method to approximate the Hessian by continuous updates throughout the training. We also show the convergence of all the methods on all the datasets in Figure~\ref{fig:loss_plot} and ~\ref{fig:acc_plot}. As observed, FedIvon exhibits slightly slower improvements in the early training phase as compared to other baselines but soon outperforms them owing to its improved Hessian approximation as training progresses. Moreover, unlike FedLaplace which fits Gaussian distribution to the client's model using Laplace approximation evaluated at MAP estimate, FedIvon approximates the Hessian over the entire course of its training, resulting in much better predictive uncertainty estimates. 
As FedIvon approximates the posterior at both the server and client, it performs well even in scenarios where clients have very limited data (fewer than 50 samples). These results are presented in the supplementary material.

\setlength{\tabcolsep}{3pt}
\begin{table}[!htbp]
\centering
\scriptsize
\begin{tabular}{ccccccccccccc}
\toprule
\multirow{2}{*}{Models} & \multicolumn{4}{c}{EMNIST} & \multicolumn{4}{c}{CIFAR-10} & \multicolumn{4}{c}{SVHN} \\
\cmidrule(lr){2-5} \cmidrule(lr){6-9} \cmidrule(lr){10-13}
 & ACC($\uparrow$) &ECE($\downarrow$) & NLL($\downarrow$) & BS($\downarrow$) & ACC($\uparrow$) & ECE($\downarrow$) & NLL($\downarrow$) & BS($\downarrow$) &  ACC($\uparrow$) & ECE($\downarrow$) & NLL($\downarrow$) & BS($\downarrow$)  \\
\midrule
FedAvg & 91.66 & 0.0405 & 0.3355 & 0.1303 & 62.25 & 0.0981 & 1.199 & 0.5191 & 82.14 & 0.0311 & 0.6857 & 0.2640 \\
FedLaplace & 91.33 & 0.0381 & 0.3255 & 0.1314 & 61.80 & 0.1072 & 1.233 & 0.5284 & 81.99 & 0.0211 & 0.6423 & 0.2627 \\
FedIvon@mean & \textbf{93.14} & 0.0349 & 0.2821 & 0.1075 & \textbf{62.92} & 0.0983 & 1.1500 & 0.5114 & 84.54 & 0.0241 & 0.5624 & 0.2256\\
FedIvon & 93.09 & \textbf{0.0188} & \textbf{0.2341} & \textbf{0.1019} & 62.54 & \textbf{0.0312} & \textbf{1.0790} & \textbf{0.5021} & \textbf{84.76} & \textbf{0.0148} & \textbf{0.5303} & \textbf{0.2210} \\
\bottomrule
\end{tabular}
\caption{Test accuracy(ACC), Expected Calibration Error (ECE), Negative Log Likelihood (NLL), and Brier Score (BS)}
\label{tab:result_table}
\vspace{0cm}
\end{table}

\subsection{Out-of-Distribution Detection Task}

Predictive uncertainty of the model plays a crucial role in uncertainty-driven tasks such as OOD detection and active learning. We evaluate FedIvon and the baselines for distinguishing OOD inputs from in-distribution inputs using their predictive uncertainty. Given any input $\mathbf{x}$, the predictive uncertainty of the model's output is given by its Shannon entropy and is used to filter OOD inputs. 
We simulate this task by randomly sampling $5000$ images from the OOD dataset and mixing it with an equal number of randomly sampled inputs from the test split of the training dataset. 

\begin{table}[!htbp]
    \centering
    \begin{tabular}{cccc}
         \hline
         Models& EMNIST & CIFAR-10 & SVHM\\
         \hline
         FedAvg& 0.8910 & \textbf{0.7896} & 0.7975\\
         FedLaplace& 0.8297  &0.7513  &0.8222 \\
         FedIvon& \textbf{0.9032}  & 0.7662 & \textbf{0.8233} \\
         \hline
    \end{tabular}
    \caption{AUROC ($\uparrow$) score for OOD/in-domain data detection}
    \label{tab:ood_tab}
\end{table}
Specifically, we use EMNIST, CIFAR-10, and SVHN as the OOD dataset for the models trained on EMNIST, SVHN, and CIFAR-10 respectively. We report the AUROC (area under the ROC curve) metric for all the methods on all the datasets in Table~\ref{tab:ood_tab} which shows that FedIvon achieves better or competitive AUROC scores as compared to the other baselines.

\subsection{Ablation Studies}
\begin{figure}
    \centering
    \includegraphics[width=1\linewidth]{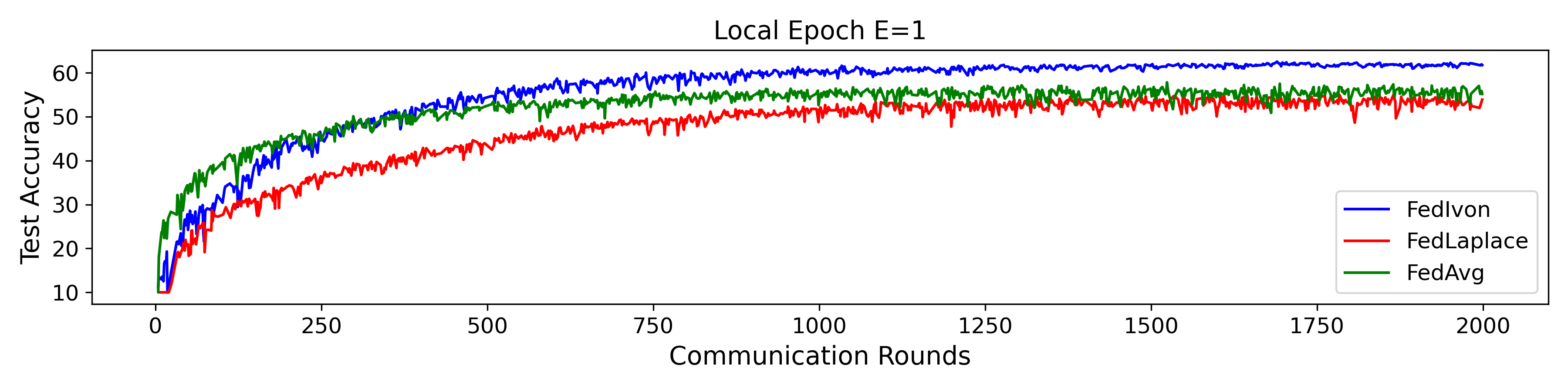}
    \includegraphics[width=1\linewidth]{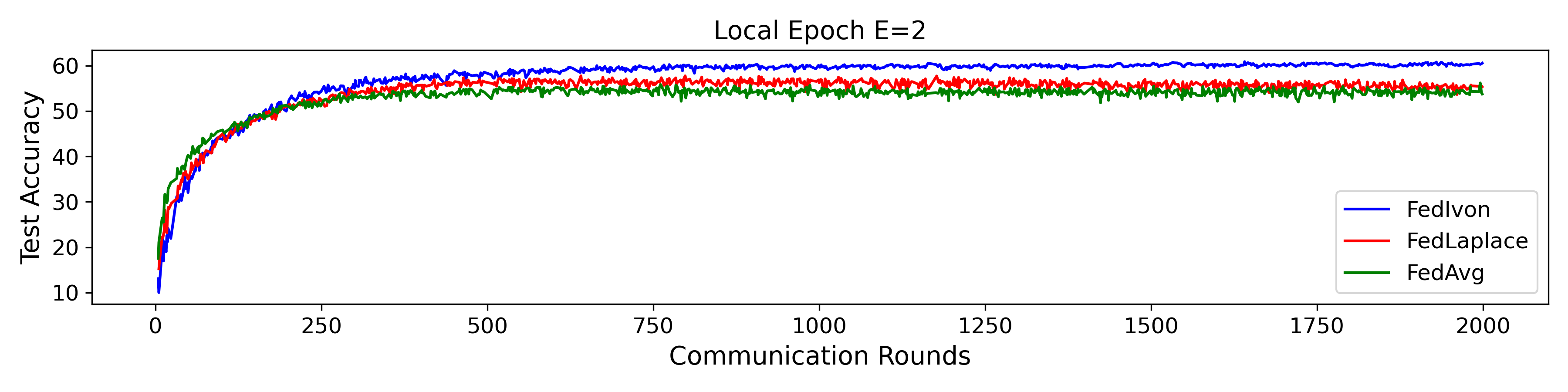}
    \caption{Convergence of all the methods on CIFAR-10 dataset with varying local epochs}
    \label{fig:ab1}
\end{figure}

In our federated learning experiments, we set $E=2$ for the number of local epochs in the client's update. In this section, we empirically investigate the impact of varying the number of local epochs on the convergence behavior of different methods in the server.
Figure \ref{fig:ab1} shows the convergence plots for varying values of $E$. When $E=1$, FedIvon shows slower convergence compared to FedAvg, and FedLaplace converges even more slowly than FedIvon. The slower convergence in FedIvon can be attributed to the way gradients are computed. Specifically, FedIvon uses stochastic sampling of the weights to estimate gradients, and at initialization, this leads to less accurate gradient estimates, which in turn causes slower convergence. Similarly, FedLaplace, which requires the calculation of a MAP estimate, also suffers from slow convergence. With only one epoch of training, the MAP estimate is suboptimal, leading to slower convergence.
When $E=2$, all methods show improved convergence compared to when $E=1$. This improvement is likely due to more training iterations allowing for better gradient and MAP estimates. In the case of FedLaplace, the MAP estimate becomes more accurate with increased training, resulting in faster convergence. However, FedIvon still outperforms both FedAvg and FedLaplace after a few communication rounds. This improvement can be attributed to the method's ability to refine gradient estimates over successive communication rounds, allowing FedIvon to overcome its initial slower convergence.

\section{Experiments: Personalized FL}

For personalized FL experiments, we focus on two types of data heterogeneity in the clients similar to \cite{zhu2023confidence} for classification task. We compare our approach \texttt{FedIvon} against personalized federated baselines (\texttt{pFedME}~\cite{t2020personalized}, \texttt{pFedBayes}~\cite{zhang2022personalized}, and \texttt{pFedVEM}~\cite{zhu2023confidence}).
\begin{itemize}
    \item \textbf{Class distribution skew}: In class distribution skew, clients have data from only a limited set of classes. To simulate this, we use the CIFAR-10 dataset and assign each client data from a random selection of 5 out of the 10 classes.
    \item \textbf{Class concept drift}: To simulate class concept drift, we use the CIFAR-100 dataset, which includes 20 superclasses, each containing 5 subclasses. For each client, we randomly select one subclass from each superclass (1 out of 5). The client’s local data is then drawn exclusively from these selected subclasses, creating a shift in label concepts across clients. We define the classification task as predicting the superclass.
\end{itemize}
To model data quantity disparity, we randomly divide the training set into partitions of varying sizes by uniformly sampling slice indices, then assign each partition to a different client. 
\subsection{Setup}
We evaluate our approach in 3 different settings: number of clients $K \in \{50, 100, 200\}$. We followed the same model architectures as the prior work \cite{zhu2023confidence}. A simple 2-convolution layered-based model is used for CIFAR-10, while a deeper model having 6 convolution layers is used for the CIFAR-100 dataset. We assess both a personalized model (PM) and a global model (GM) at the server. The PMs are evaluated using test data that matches the labels (for label distribution skew) or subclasses (for label concept drift) specific to each client, while the GM is evaluated on the entire test set. All experiments are repeated 3 times, using the same set of 3 random seeds for data generation, parameter initialization, and client sampling. The results are presented in the Table \ref{tab:pFL}.
\begin{table}[t]
\small
    \centering
    \begin{tabular}{ccc}
        \toprule
        params & CIFAR-10 & CIFAR-100 \\
        \midrule
         initial learning rate& 0.1 & 0.1\\
         final learning rate& 0.001 & 0.001 \\
         weight decay &  1e-3 & 1e-3 \\
         batch size  & 32 & 32\\
         ESS ($\lambda$) & 10000 & 10000\\
         initial hessian ($h_0$) & 1.0 & 1.0\\
         MC sample while training &1  & 1\\
         MC samples while test  & 64 & 64\\
         \bottomrule
    \end{tabular}
    \caption{Ivon Hyperparameters for personalized FL experiments}
    \label{tab:my_tab2}
\end{table}

\begin{table}[!htbp]
    \centering
    \scriptsize
    \begin{tabular}{cccccccc}
    \toprule \multirow{2}{*}{Dataset} & \multirow{2}{*}{Method} & \multicolumn{2}{c}{50 Clients} & \multicolumn{2}{c}{100 Clients} & \multicolumn{2}{c}{200 Clients} \\
    \cmidrule(lr){3-4} \cmidrule(lr){5-6} \cmidrule(lr){7-8}
    & & PM & GM & PM & GM & PM & GM \\
    \midrule \multirow{5}{*}{CIFAR10} & Local & $56.9 \pm 0.1$ & - & $52.1 \pm 0.1$ & - & $46.6 \pm 0.1$ & - \\

     & pFedME~\cite{t2020personalized} & $72.3 \pm 0.1$ & $56.6 \pm 1.0$ & $71.4 \pm 0.2$ & $6 0 . 1 \pm 0.3$ & $68.5 \pm 0.2$ & $58.7 \pm 0.2$ \\
     & pFedBayes~\cite{zhang2022personalized} & $71.4 \pm 0.3$ & $52.0 \pm 1.0$ & $68.5 \pm 0.3$ & $53.2 \pm 0.7$ & $64.6 \pm 0.2$ & $51.4 \pm 0.3$ \\
     & pFedVEM~\cite{zhu2023confidence} & $7 3 . 2 \pm 0 . 2$ & $56.0 \pm 0.4$ & \underline{$7 1 . 9 \pm 0 . 1$ }& $6 0 . 1 \pm 0 . 2$ & \underline{$7 0 . 1 \pm 0 . 3$ }& $59.4 \pm 0.3$ \\
     & \textbf{FedIvon@mean} & \underline{$74.4 \pm 0.3$} & \underline{$67.1 \pm 1.0$} & $71.7 \pm 0.3$ & \underline{$68.4 \pm 0.2$} & $69.7 \pm 0.7$ & \underline{$68.2 \pm 0.3$} \\
     & \textbf{FedIvon} & $\mathbf{75.5 \pm 0.4}$ & $\mathbf{67.8 \pm 1.6}$ & $\mathbf{72.6 \pm 0.2}$ & $\mathbf{69.2 \pm 0.2}$ & $\mathbf{70.8 \pm 0.4}$ & $\mathbf{68.7 \pm 0.3}$ \\
    \midrule \multirow{5}{*}{CIFAR100} & Local & $34.3 \pm 0.2$ & - & $27.6 \pm 0.3$ & - & $22.2 \pm 0.2$ & - \\
     & pFedME~\cite{t2020personalized} & $52.5 \pm 0.5$ & $47.9 \pm 0.5$ & $47.6 \pm 0.5$ & $45.1 \pm 0.3$ & $41.6 \pm 1.8$ & $41.5 \pm 1.6$ \\
     & pFedBayes~\cite{zhang2022personalized} & $49.6 \pm 0.3$ & $42.5 \pm 0.5$ & $46.5 \pm 0.2$ & $41.3 \pm 0.3$ & $40.1 \pm 0.3$ & $37.4 \pm 0.3$ \\
     & pFedVEM~\cite{zhu2023confidence} & $6 1 . 0 \pm 0 . 4$ & $5 2 . 8 \pm 0 . 4$ & $5 6 . 2 \pm 0 . 4$ & $5 2 . 3 \pm 0 . 4$ & $5 1 . 1 \pm 0 . 6$ & $49.2 \pm 0.5$ \\
     & \textbf{FedIvon@mean} & $\underline{65.4 \pm 0.7}$ & $\underline{63.2 \pm 0.5}$ & $\underline{63.2 \pm 0.5}$ & $\underline{62.1 \pm 0.5}$ & $\underline{56.1 \pm 0.6}$ & $\underline{55.5 \pm 0.6}$\\
     & \textbf{FedIvon} & $\mathbf{66.7 \pm 0.8}$ & $\mathbf{63.8 \pm 0.7}$ & $\mathbf{63.5 \pm 0.6}$ & $\mathbf{62.4 \pm 0.6}$ & $\mathbf{56.5 \pm 0.5}$ & $\mathbf{55.7 \pm 0.7}$\\
    \bottomrule
    \end{tabular}
    \caption{Comparison of Personalized FL Methods}
    \label{tab:pFL}
\end{table}

\subsection{Results}
Table \ref{tab:pFL} presents results on CIFAR-10 and CIFAR-100 datasets, which are used to simulate different types of data heterogeneity in federated learning: CIFAR-10 models class distribution skew, where each client has data from a limited set of classes, while CIFAR-100 represents class concept drift, where each client has data from distinct subclasses within superclasses. For both datasets, we evaluate client’s average accuracy (personalized model) and server accuracy (global model) across varying client counts (50, 100, and 200).
FedIvon uses 64 Monte Carlo samples to perform Monte Carlo averaging. On the other hand, FedIvon@mean uses a point estimate using mode of the posterior.

On CIFAR-10, FedIvon achieves similar client accuracy to pFedVEM, indicating both methods perform well under class distribution skew for individual clients. However, in server accuracy, FedIvon shows a notable improvement over pFedVEM and other methods, highlighting FedIvon’s strength in aggregating data from heterogeneous clients into an accurate global model.

On CIFAR-100, which represents class concept drift, FedIvon demonstrates significant improvements over all other methods in both client’s average accuracy and server accuracy. This performance advantage in both personalized and global evaluations suggests that FedIvon is well-suited to handling concept drift, achieving higher accuracy for individual clients and in the global model. Overall, FedIvon consistently outperforms other methods, particularly in server accuracy on CIFAR-10 and in both accuracy metrics on CIFAR-100, underscoring its robustness across different data heterogeneity scenarios.

\section{Conclusion}

We presented a new Bayesian Federated Learning (FL) method that reduces the computational and communication overhead typically associated with Bayesian approaches. Our method uses an efficient second-order optimization technique for variational inference, achieving computational efficiency similar to first-order methods like Adam while still providing the benefits of Bayesian FL, such as uncertainty estimation and model personalization. We showed that our approach improves predictive accuracy and uncertainty estimates compared to both Bayesian and non-Bayesian FL methods. Additionally, our method naturally supports personalized FL by allowing clients to use the server's posterior as a prior for learning their own models. 




{\small 
\bibliographystyle{unsrtnat}
\bibliography{fedivon}
}

\newpage 

\appendices 
\section{More details on IVON}
\label{sec:ivon_details}
Computing exact gradients in equation 4 and \ref{eq:update_eq} is difficult due to the expectation term in $\mathcal{L}_k(q)$. A na\"ive way to optimize is to use stochastic gradient estimators. However, these approaches are not very scalable due to the high variance in the gradient estimates. 
Using natural gradients, \citet{khan2017conjugatecomputationvariationalinference} gave improved gradient based update equations for the variational parameters and they call this approach Natural Gradient VI (NGVI). The major difference between NVGI and original update equations is that learning rate is now adapted by the variance $\bs{\sigma_k^{t+1}}$ which makes these updates similar to Adam. 
\[
\begin{split}
    \textbf{NVGI: }
    \bs{m}^{t+1}_k &= \bs{m}^{t}_k + \beta^t {\bs{\sigma}_k^2}^{t+1} \odot [\Hat{\nabla}_{\bs{m}_k} \mathcal{L}_k(q)] \\
    {\bs{\sigma}_k^{-2}}^{t+1} &= {\bs{\sigma}_k^{-2}}^{t} - 2 \beta^t [\Hat{\nabla}_{\bs{\sigma}^2_k} \mathcal{L}_k(q)]
\end{split}
\]
Further, \citet{pmlr-v80-khan18a} showed that the NVGI update equations can be written in terms of scholastic gradient and Hessian of $\btheta$, where ${\bs{\sigma^2_k}}^t = [N(\bs{h}^t_k + \lambda)]^{-1}$. The vector  $\bs{h}^t_k$ contains an online estimate of diagonal Hessian. This approach called Variational Online Newton (VON) is similar to NGVI except that it does not require the gradients of the variational objective.

\[
\begin{split}   
    \textbf{VON: }
    \bs{m}^{t+1}_k &= \bs{m}^{t}_k - \beta^t \frac{\hat{\bs{g}}(\btheta^t) + \lambda \bs{m}^{t}_k}{\bs{h}^{t+1}_k + \lambda} \\
    \bs{h}_k^{t+1} &= (1-\beta^t)\bs{h}_k^{t} +\beta^t \text{diag}[\Hat{\nabla}^2_{\btheta \btheta} \bar{\ell}_k(\btheta^t)]
\end{split}
\]
In the update of VON  for non-convex objective functions, the Hessian can be negative which might make $\bs{\sigma_k^t}$ negative, and break VON. To mitigate this issue \citet{pmlr-v80-khan18a} used a Generalized Gauss-Newton (GGN) approximation of Hessian which is always positive. This method is called VOGN.
$$
\nabla_{\theta_j \theta_j}^2 \bar{\ell}_k(\btheta^t) \approx \frac{1}{M} \sum_{i \in \mathcal{M}}\left[\nabla_{\theta_j} {\ell}^i_k(\btheta^t)\right]^2:=\hat{h}_j(\boldsymbol{\theta})
$$
\[
\begin{split}
    \textbf{VOGN: }
    \bs{m}^{t+1}_k &= \bs{m}^{t}_k - \beta^t \frac{\hat{\bs{g}}(\btheta^t) + \lambda \bs{m}^{t}_k}{\bs{h}^{t+1}_k + \lambda} \\
    \bs{h}_k^{t+1} &= (1-\beta^t)\bs{h}_k^{t} +\beta^t \hat{h}_j(\boldsymbol{\theta}^t)
\end{split}
\]
VOGN \cite{pmlr-v80-khan18a} improves these equations where Gauss Newton estimation is used instead of Hessian which gives similar update equations as the Adam optimizer.
However, it still uses per-sample squaring which is costly as compared to Adam. 
\[
\begin{split}
    \textbf{IVON: }
    \widehat{\mathbf{h}}_k^t &= \widehat{\nabla} \bar{\ell}_k(\boldsymbol{\theta}) \cdot \frac{\boldsymbol{\theta}-\mathbf{m}_k^t}{\boldsymbol \sigma_k^{2^t}}\\
    \mathbf{h}^{t+1}_k &= (1-\rho) \mathbf{h}_k^{t}+\rho \widehat{\mathbf{h}}_k^t+\frac{1}{2} \rho^2\frac{(\mathbf{h}_k^t-\widehat{\mathbf{h}}_k^t)^2} {\left(\mathbf{h}_k^t+s_0 / \lambda\right)}
\end{split}
\]

Further, \citet{shen2024variationallearningeffectivelarge} improved these update equations and provided much more efficient update equations similar to Adam optimizer, which is essentially the improved variational online Newton (IVON) algorithm~\cite{shen2024variationallearningeffectivelarge}.

\section{Reliability diagrams for FL experiments}
\begin{figure*}
        \centering
        \includegraphics[scale=0.45]{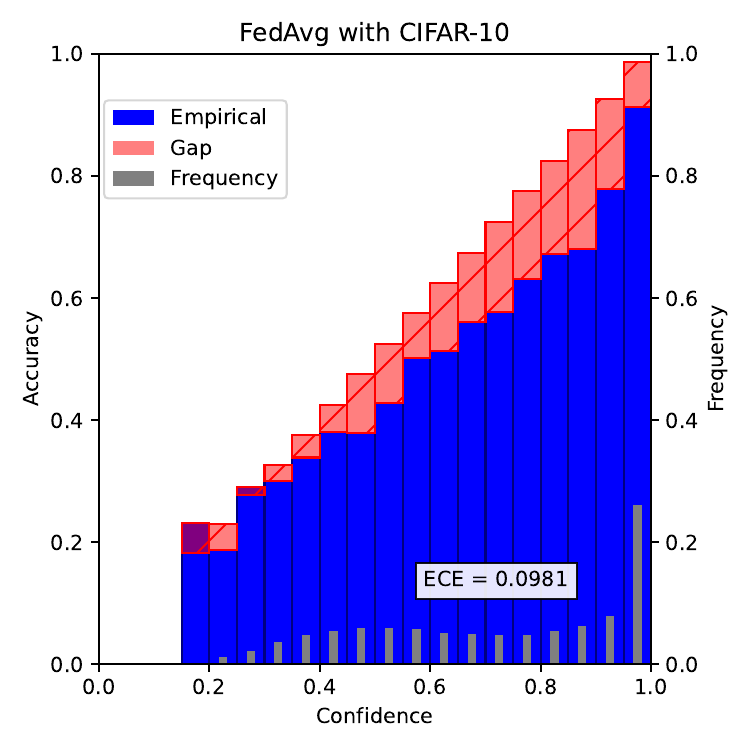}
        \includegraphics[scale=0.45]{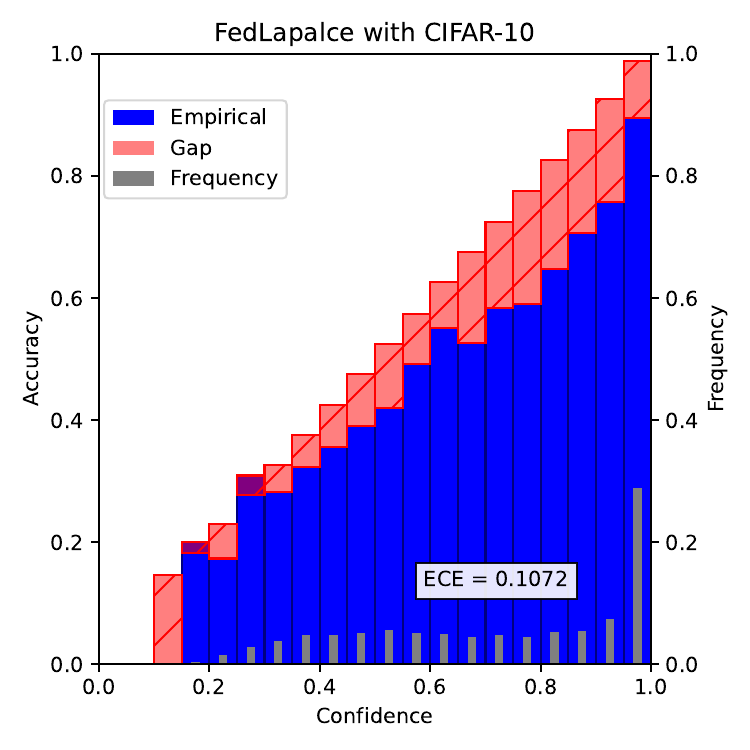}
        \includegraphics[scale=0.45]{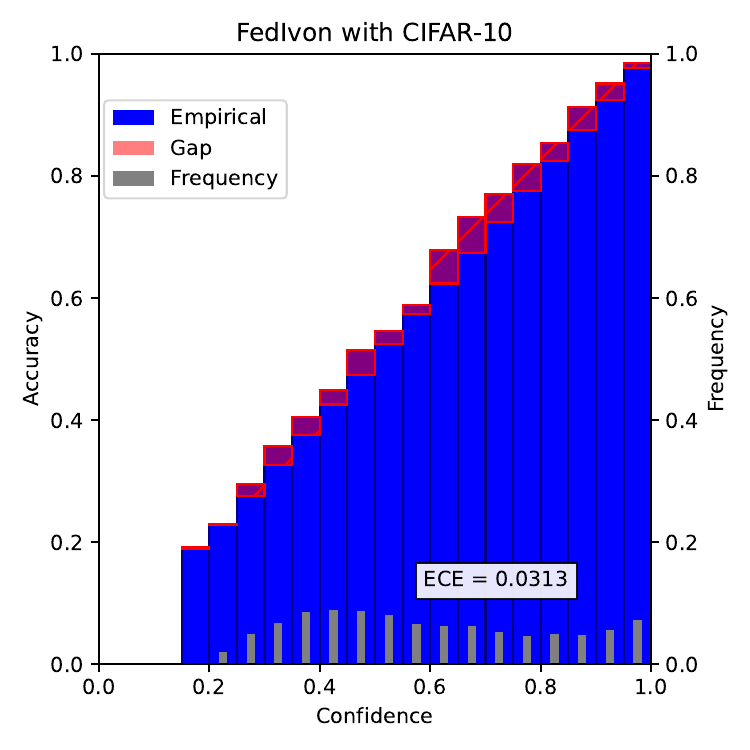}
    \caption{Reliability diagrams for CIFAR-10 experiments (left: FedAvg, center: Fedlaplace, right: FedIvon).}
    \label{fig:rbl1}
\end{figure*}

\begin{figure*}
        \centering
        \includegraphics[scale=0.45]{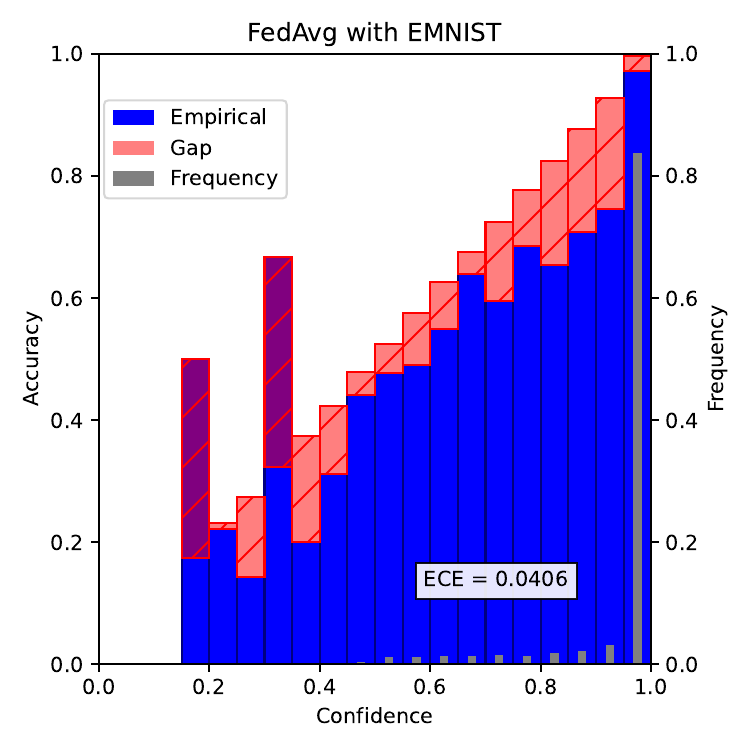}
        \includegraphics[scale=0.45]{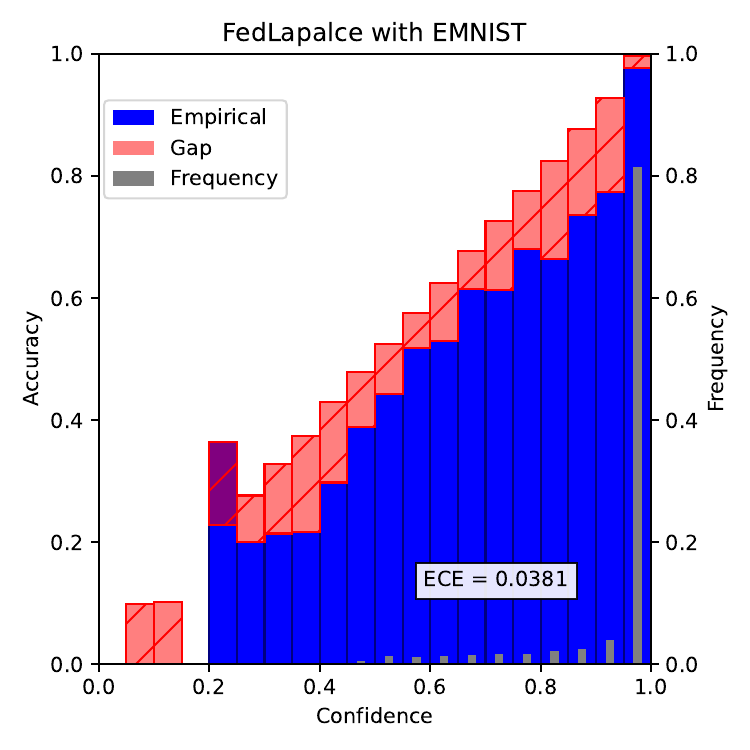}
        \includegraphics[scale=0.45]{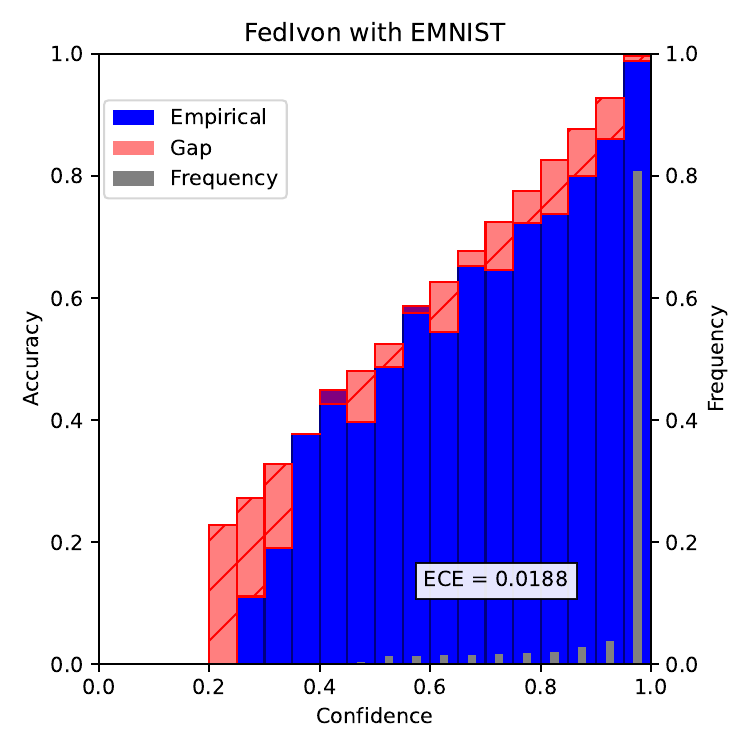}
    \caption{Reliability diagrams for EMNIST experiments (left: FedAvg, center: Fedlaplace, right: FedIvon).}
    \label{fig:rbl2}
\end{figure*}

Figures \ref{fig:rbl1} and \ref{fig:rbl2} show the reliability diagrams for CIFAR-10 and EMNIST experiments, respectively. The diagrams indicate that Fedivon has better-calibrated predictions compared to FedAvg and FedLaplace, as shown by its lower Expected Calibration Error (ECE).

\section{Client data distribution in FL experiments}
\begin{figure*}
    \centering
    \includegraphics[scale=0.30]{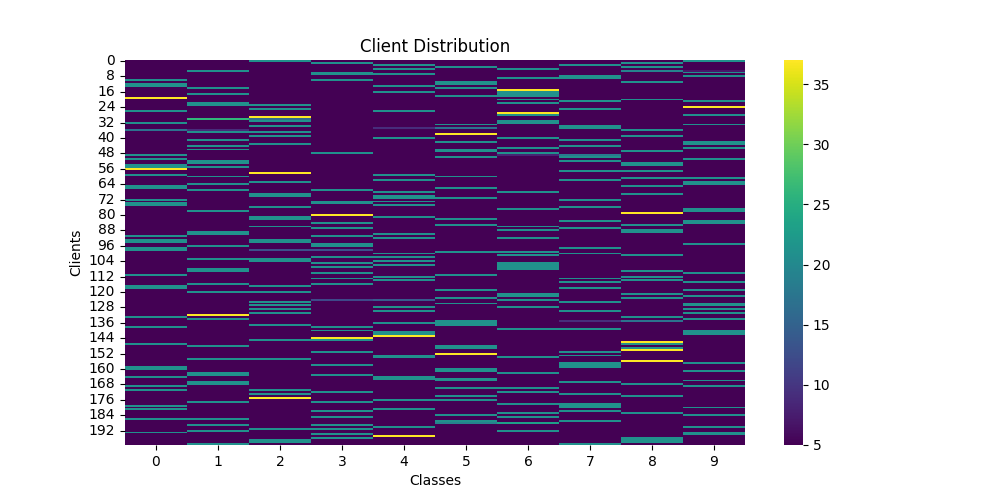}
    \includegraphics[scale=0.30]{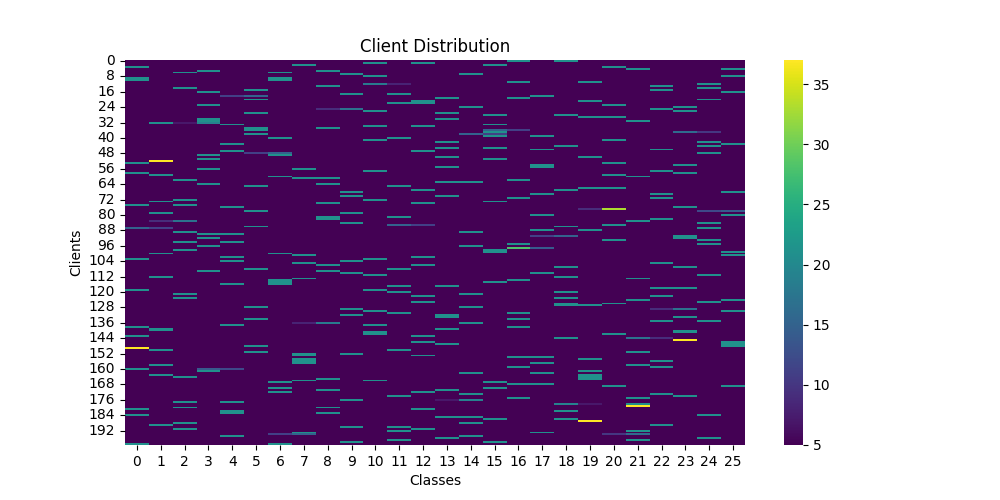}
    \includegraphics[scale=0.30]{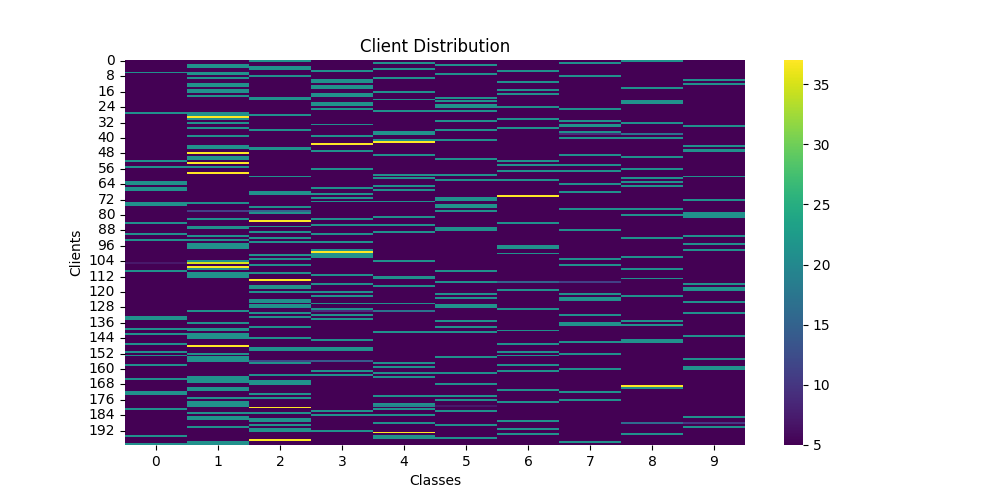}
    \caption{Client data distribution for CIFAR-10, EMNIST, and SVHN dataset used in FL experiments. }
    \label{fig:dist3}
\end{figure*}

Figure \ref{fig:dist3} illustrates the data distribution among clients used in the  FL experiments. Each client has a highly imbalanced dataset, with the number of samples per client ranging from 5 to 32. Additionally, each client’s dataset is limited to only a subset of classes, further emphasizing the non-IID nature of the data. This experimental setup poses significant challenges for training a robust global server model, as the limited and biased data from individual clients must be aggregated effectively to learn a model capable of generalizing across all classes. This scenario highlights the complexities and practical relevance of federated learning in real-world applications.

\section{Client data distribution in pFL }
\begin{figure*}
    \centering
    \includegraphics[scale=0.30]{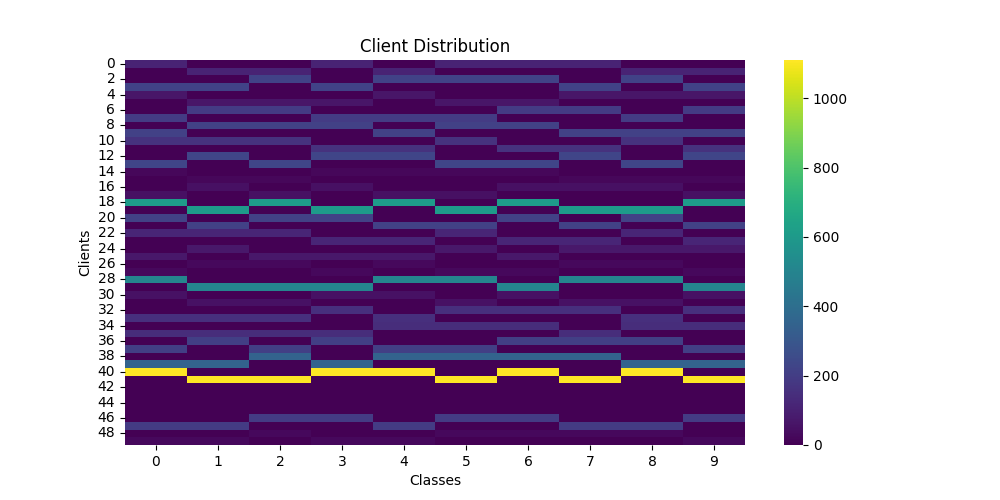}
    \includegraphics[scale=0.30]{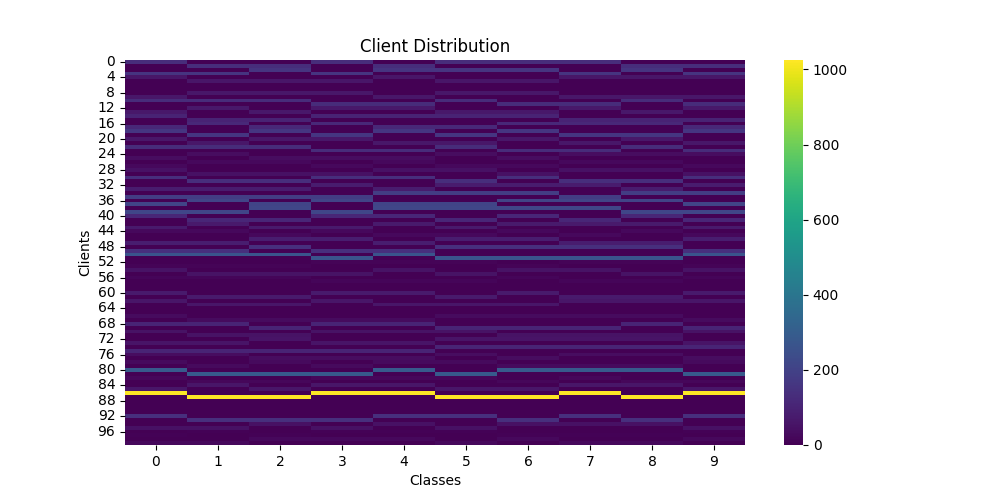}
    \includegraphics[scale=0.30]{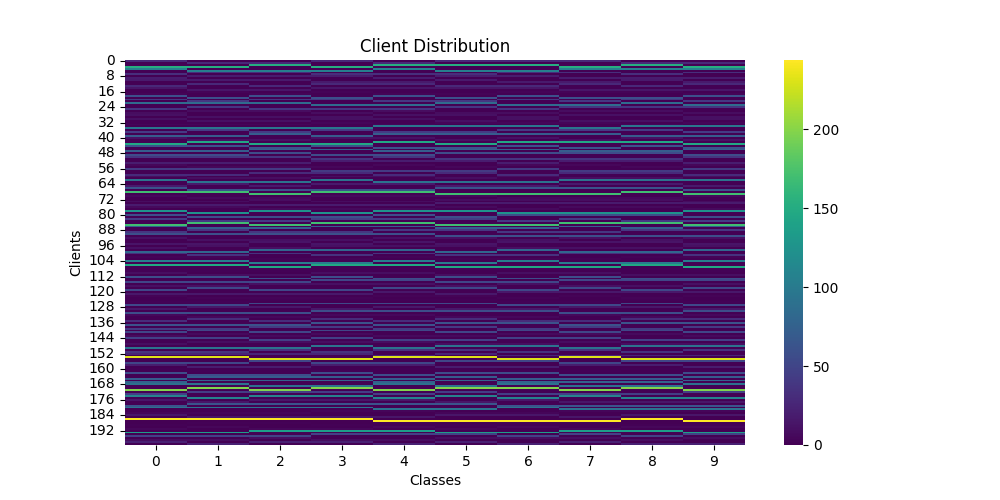}
    \caption{Client data distribution for CIFAR-10 dataset used in pFL experiments (left: 50 clients, right: 100 clients, bottom: 200 clients).}
    \label{fig:dist1}
\end{figure*}

\begin{figure*}
    \centering
    \includegraphics[scale=0.30]{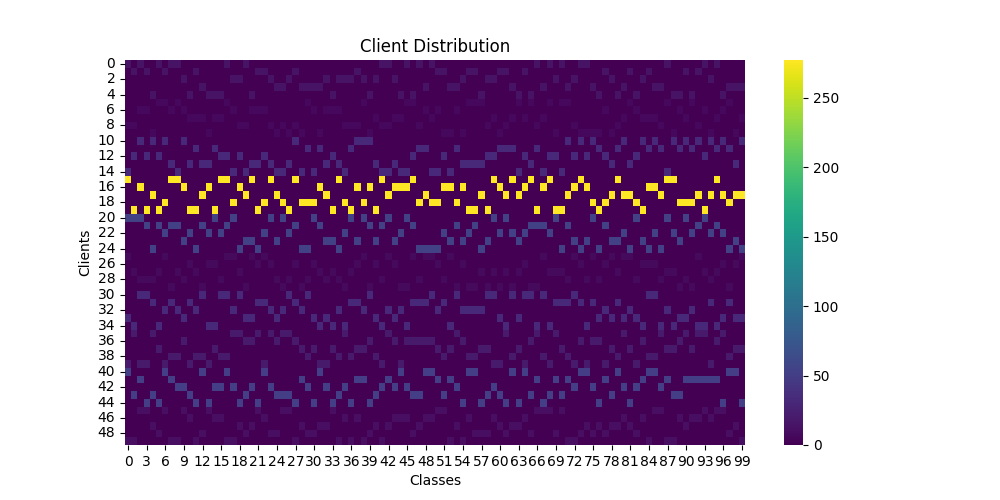}
    \includegraphics[scale=0.30]{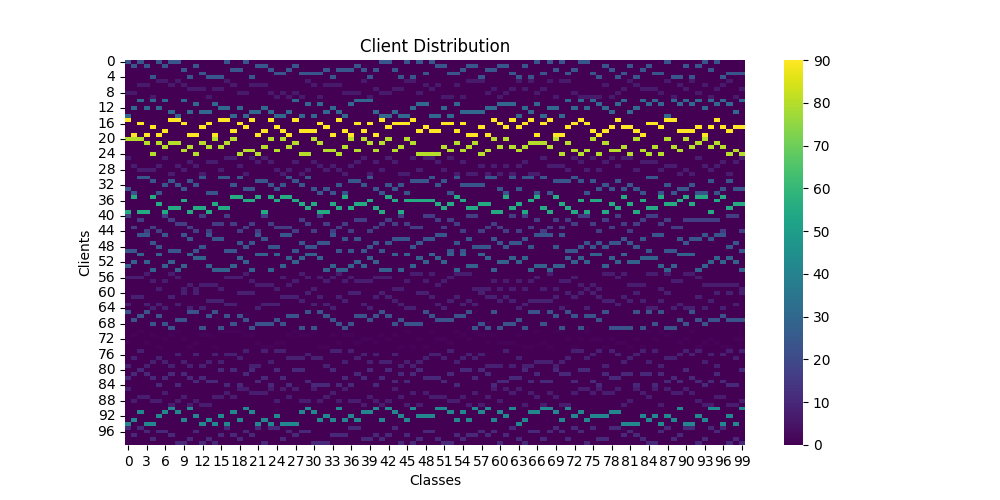}
    \includegraphics[scale=0.30]{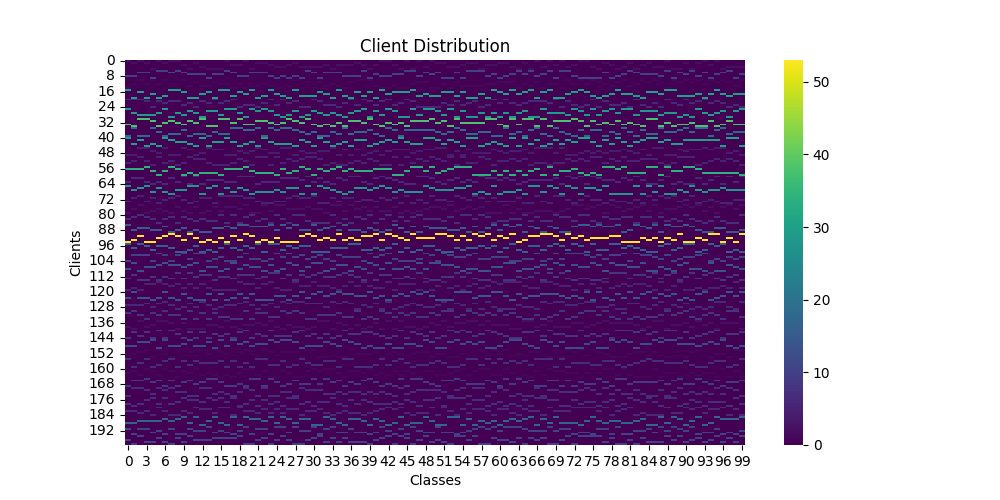}
    \caption{Client data distribution for CIFAR-100 dataset used in pFL experiments (left: 50 clients, right: 100 clients, bottom: 200 clients).}
    \label{fig:dist2}
\end{figure*}

Figure \ref{fig:dist1} illustrates the distribution of data points across classes and clients in three pFL experimental setups with 50, 100, and 200 clients. The number of data points per client varies significantly, with some clients having over 1,000 data points and others fewer than 5, indicating a high degree of imbalance. Despite this, every client retains examples from most classes, which is crucial for training personalized models that adapt to the unique data distribution of each client. This setup highlights the challenge of learning effective personalized models in pFL. Similarly, Figure \ref{fig:dist2} shows the data distribution for the CIFAR-100 dataset.

\end{document}